\documentclass[11pt]{article}

\usepackage[final]{acl}

\usepackage{times}
\usepackage{latexsym}

\usepackage[T1]{fontenc}

\usepackage[utf8]{inputenc}

\usepackage{microtype}

\usepackage{inconsolata}

\usepackage{graphicx}

\usepackage{amssymb}
\usepackage{amsmath}
\usepackage{array}
\usepackage{booktabs}
\usepackage{multirow}
\usepackage{tcolorbox}
\tcbuselibrary{skins, breakable, theorems}
\usepackage{listings}

\newcolumntype{L}[1]{>{\raggedright\arraybackslash}p{#1}}

\title{TimelineAgent: An LLM-based Framework for Query-Centric Event Timeline Extraction Across Multiple Documents}

\author{
  \textbf{Qingkai Min\textsuperscript{1,2}}\quad
  \textbf{Yunxin Bao\textsuperscript{1}}\quad
  \textbf{Yifeng Lu\textsuperscript{2,3}}\quad
  \textbf{Zhenyu Feng\textsuperscript{4}}
\\
  \textbf{Yulong Chen\textsuperscript{5,6}}\quad
  \textbf{Andreas Vlachos\textsuperscript{5}}\quad
  \textbf{Yue Zhang\textsuperscript{2,3}\thanks{Corresponding author}}
\\
  \textsuperscript{1}Zhejiang University
  \quad
  \textsuperscript{2}School of Engineering, Westlake University
\\
  \textsuperscript{3}Institute of Advanced Technology, Westlake Institute for Advanced Study
\\
  \textsuperscript{4}Universität des Saarlandes
  \quad
  \textsuperscript{5}University of Cambridge
  \quad
  \textsuperscript{6}University of Aberdeen
\\
  \texttt{\{minqingkai, zhangyue\}@westlake.edu.cn}
}

\begin{document}
\maketitle
\begin{abstract}
Query-centric cross-document event timeline extraction aims to construct event timelines from multiple documents under diverse query focuses. The task is very challenging due to the heterogeneous event modeling and reasoning requirements arising from both the intrinsic complexity of cross-document timeline construction and the diversity of query focuses.
In this paper, we propose \emph{TimelineAgent}, an LLM-based multi-agent framework featuring a multi-stage reasoning workflow, where heterogeneous event features are organized into reusable skills and orchestrated through a goal-oriented protocol.
To facilitate evaluation, we construct a benchmark covering diverse query focuses. Experimental results show that TimelineAgent consistently outperforms existing methods, offering a practical framework for complex real-world timeline construction.
\end{abstract}

\section{Introduction}

\emph{Events} are fundamental units representing occurrences or changes within a given context and a central target of information extraction~\cite{10.5555/645856.669801}. While individual events convey isolated facts, organizing them chronologically reveals their evolution and interactions over time~\cite{pustejovsky2003timeml}. An \emph{event timeline} is a structured representation that orders events temporally to provide a coherent view of their progression~\cite{do-etal-2012-joint}. Across domains, professionals rely on event timelines to access reliable information, interpret complex narratives, and make informed decisions. Such timelines are widely used in fields such as news analysis, crisis monitoring, and fact checking~\cite{wang-etal-2015-socially,rajaby-faghihi-etal-2022-crisisltlsum,chen-etal-2025-pledgetracker}. For example, crisis response teams use event timelines to monitor the evolution of emergencies and coordinate timely responses.

\begin{figure}[t]
    \centering
    \includegraphics[width=1\linewidth]{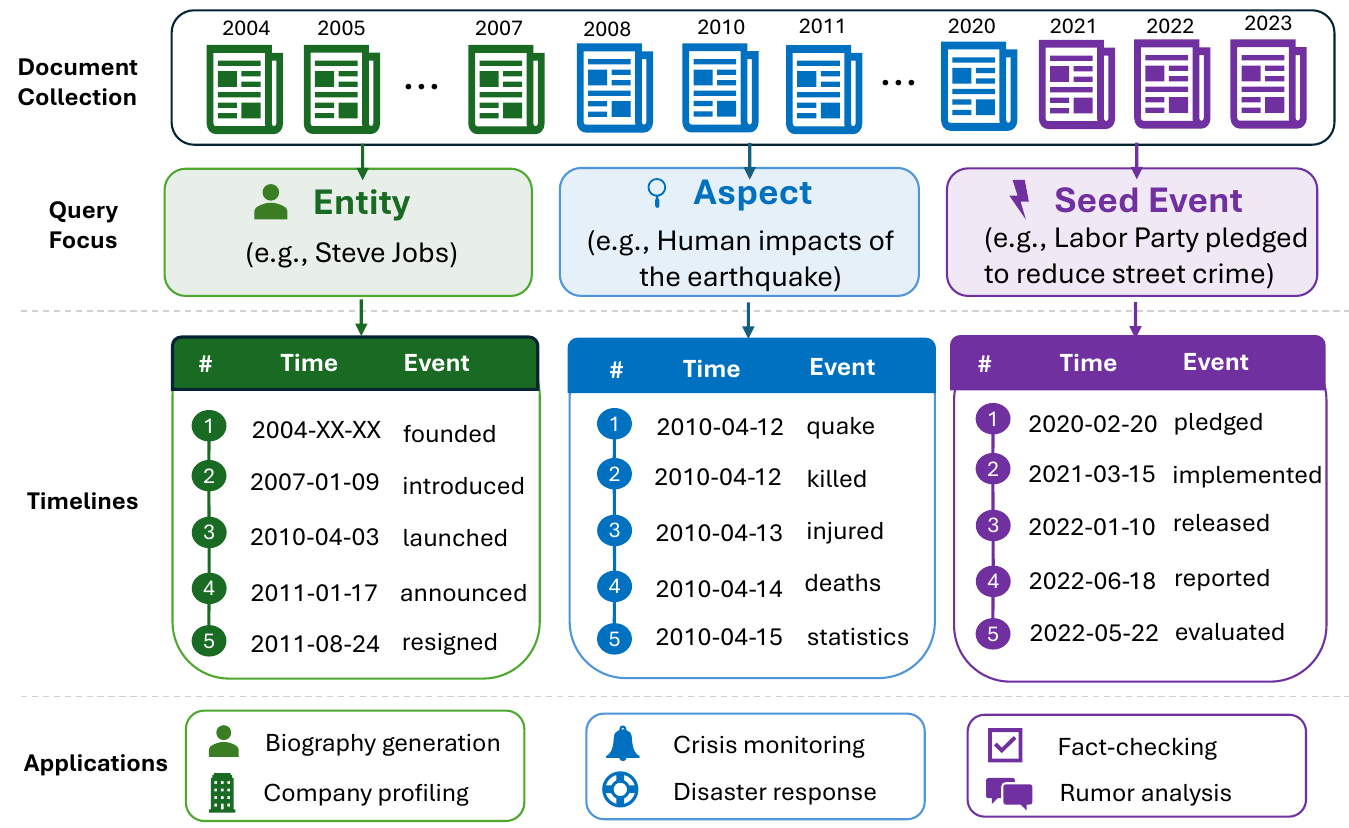}
    \caption{Examples of query-centric cross-document event timelines under different query focuses. Entity queries target event arguments (participants), Aspect queries characterize event types, and Seed Event queries capture event-to-event relations, together forming the basis for event understanding.
    }
    \label{fig:timeline}
\end{figure}

In many real-world applications, users require \emph{query-centric timelines} aggregated from multiple documents to satisfy diverse information needs. As illustrated in Figure~\ref{fig:timeline}, within news articles published between 2004 and 2023, users may focus on events involving specific entities, such as \textit{Steve Jobs}, to track his evolving role in the company. Constructing such a timeline requires a comprehensive event-centric information extraction pipeline. Starting with event detection~\cite{yue-etal-2023-zero} to identify candidate events, relevance matching selects query-relevant events by first extracting event arguments~\cite{yang-etal-2023-shot}, and then leveraging entity coreference~\cite{khosla-etal-2021-evaluating} and bridging anaphora~\cite{hou-etal-2014-rule} to capture both direct and indirect associations between query entities and event arguments.

In addition, timeline events must satisfy general eligibility constraints, such as factuality (i.e., events not actually occurring should not be included in the timeline)~\cite{Saur2009FactBankAC} and temporality (i.e., events must possess a temporal attribute)~\cite{ning-etal-2018-multi}. Constructing coherent cross-document timelines also requires event temporal anchoring~\cite{wang-zhao-2024-tram} and cross-document event coreference resolution~\cite{yu-etal-2022-pairwise}. Temporal anchoring demands both temporal understanding~\cite{wang-etal-2022-dct} and reasoning capabilities~\cite{chu2024timebench}, while event coreference requires compatibility in both event semantics and arguments~\cite{huang-etal-2019-improving}.

In particular, the relevance matching requirements vary across query focuses. As shown in Figure~\ref{fig:timeline}, aspect-focused queries, such as \textit{Human impacts of the earthquake}, require identifying fine-grained event semantics~\cite{liu-etal-2021-fine} related to casualties or fatalities, as well as reasoning over preceding events that led to these impacts~\cite{gao-etal-2019-modeling}. Seed-event-focused queries, such as \textit{The Labour Party pledged to reduce street crime}, involve tracking downstream events triggered by the seed, such as policy changes or subsequent actions. These events must be semantically and causally related to the seed and constrained by event arguments, including participants or affected targets~\cite{chen-etal-2025-pledgetracker}.

Building on the timeline construction processes for different queries described above, we formulate the task of \emph{Query-centric Cross-document Event Timeline Extraction (QUEST)}. To address the complexity of timeline construction and query diversity, QUEST requires modeling heterogeneous event features, including syntactic, semantic, argument, temporal, causal, and factuality.
For example, semantic features include predefined event types for aspect-focused queries and contextual semantic similarity for seed-event-focused queries. Argument features for entity-focused queries rely on PropBank-based semantic frames, while those for seed-event-focused queries require both FrameNet-based and predefined semantic schemas.

We propose \emph{TimelineAgent}, an LLM-based multi-agent framework that separates task-agnostic event modeling from the query- and stage-specific reasoning required for timeline construction.
Specifically, we establish a \emph{multi-stage workflow} (as shown in Table~\ref{tab:feature_cat}), where each stage is handled by a dedicated agent performing reasoning toward a clear sub-objective. For example, relevance matching  combines semantic and argument features for query-aware event matching, while temporal anchoring first extracts temporal signals and then performs temporal inference to derive normalized timestamps.
Heterogeneous event features are systematically categorized and decoupled into a \emph{skill library of reusable atomic operations}, reducing interference among different reasoning processes. For example, temporal signals are decomposed into atomic skills such as tense analysis, aspect analysis, and temporal expression extraction, while argument modeling supports multiple schema configurations including predefined schemas, PropBank, and FrameNet.
To coordinate the agent’s invocation of skills, we design a \emph{goal-oriented orchestration protocol} that composes and configures appropriate skills according to specific reasoning substeps and query requirements.
For example, the protocol may combine multiple temporal skills to support temporal signal collection, or configure the argument extraction skill with different schemas under different query settings.
Table~\ref{tab:feature_cat} provides an overview of representative skills required at each stage.

To facilitate evaluation, we construct a new benchmark consisting of existing datasets and newly extended annotated data. For entity-focused queries, we use the SemEval-2015 Task 4 dataset~\cite{minard-etal-2015-semeval}. For aspect- and seed-event-focused queries, we adapt existing datasets—EventStoryLine~\cite{caselli-vossen-2017-event} and PledgeTracker~\cite{chen-etal-2025-pledgetracker}—through manual annotation and reformulation to align with QUEST. Experimental results show that TimelineAgent consistently outperforms both SFT and ICL LLM baselines, as well as the task-agnostic multi-agent framework TDAG~\cite{Wang2024TDAGAM}. Further analysis demonstrates that each component of the framework contributes substantially to overall performance.

\begin{table}[t]
\centering
\resizebox{\columnwidth}{!}{%
\begin{tabular}{ll}
\hline
\textbf{Workflow Stage} & \textbf{Skill Categories (No. of Skills)}                 \\ \hline
Event Detection    & Syntactic(Parsing(3)+Resource(1))             \\
Relevance Matching      & Semantic(2)$^{2,3}$, Argument(4)$^{1,3}$, Causal(1)$^{2,3}$   \\
Temporal Anchoring & Temporal(Extract(4)+Inference(3))              \\
Eligibility Filtering  & Factuality(4)$^{1,2,3}$, Semantic(1)$^{3}$, Temporal(1)$^{3}$ \\
Event Coreference  & Semantic(1), Argument(1) \\ \hline
\end{tabular}%
}
\caption{Stage-wise coupling of skill categories and query types: Entity$^{1}$, Aspect$^{2}$, and Seed-event$^{3}$. The specific skills are listed in Table~\ref{tab:skills} (Appendix~\ref{sec:appendix2.1}).}
\label{tab:feature_cat}
\end{table}

To the best of our knowledge, we are the first to investigate cross-document event timeline extraction through a unified LLM-based agent framework across multiple query formulations.\footnote{The code and dataset are publicly available at \href{https://github.com/taolusi/QUEST}{https://github.com/taolusi/QUEST}.}

\section{Related Work} \paragraph{Event Timeline Extraction} Document-level tasks primarily focus on modeling temporal relations among events~\cite{chambers-etal-2014-dense,ning-etal-2018-multi,alsayyahi-batista-navarro-2023-timeline}. In contrast, constructing timelines across documents often requires query-specific information aggregation. Prior work has proposed building timelines centered on entities~\cite{ji-etal-2009-cross,minard-etal-2015-semeval,laparra2017multi,yao-etal-2024-overview}. \citet{fang2021cross} and \citet{10.1145/3543507.3583295} use loosely defined criteria, such as event similarity or event salience.
Methodologically, early work relied on hand-crafted modules to build pipelines, such as NER and temporal clustering, which lack robust semantic understanding~\cite{Moulahi2015HeidelToulAB,Laparra2015DocumentLT,NavarroColorado2016CrossdocumentEO,piskorski2020timelines}. More recently, end-to-end supervised fine-tuning approaches have been applied in certain domains, such as cancer treatment~\cite{zhang2025uw,jain-etal-2025-team}.

\paragraph{LLM-based Agent Frameworks for Complex IE}
Recent advances in LLMs have promoted the development of agent-based frameworks for complex IE tasks through task decomposition, rich feature modeling and reasoning.
\citet{gangi-reddy-etal-2025-infogent} proposed a framework consisting of a navigator, an extractor, and an aggregator for information extraction across complex web environments. \citet{jiao-etal-2024-text2db} introduced frameworks with components such as observers, planners, and analyzers to support extraction over complex database schemas. \citet{li-etal-2025-aid} proposed a framework consisting of an executor, a tool pool, and a validator for information extraction and integration from documents with complex layouts and cross-references.

Our work differs from existing studies in two main aspects. First, prior work on event timeline extraction typically focuses on a single query type, whereas QUEST requires handling diverse query focuses for more realistic applications. Second, existing methods mainly rely on either hand-crafted pipelines or supervised end-to-end modeling. In contrast, our framework leverages LLM-based agents without task-specific fine-tuning, enabling richer event modeling and better generalization across query settings. More broadly, following recent efforts on designing task-specific architectures for complex information extraction, we develop a multi-agent framework tailored to complex cross-document event reasoning in QUEST.

\section{Task Definition}

Given a document collection $D = \{D_1, D_2, \dots, D_n\}$ and a target query $q$, the task of QUEST is to construct a temporally ordered timeline consisting of query-relevant events:
\[
\mathcal{L}_q = \{(\tau_i, E_i)\}_{i=1}^{m_q}
\]
where each tuple $(\tau_i, E_i)$ denotes a query-relevant timeline event entry. $\tau_i$ denotes the normalized event date (YYYY-MM-DD), following the TimeML specification~\cite{pustejovsky-etal-2010-iso}, where unknown temporal components are represented using ``X''. $E_i$ denotes a set of coreferential event mentions consolidated across documents.

\begin{figure*}[t]
    \centering
    \includegraphics[width=1\linewidth]{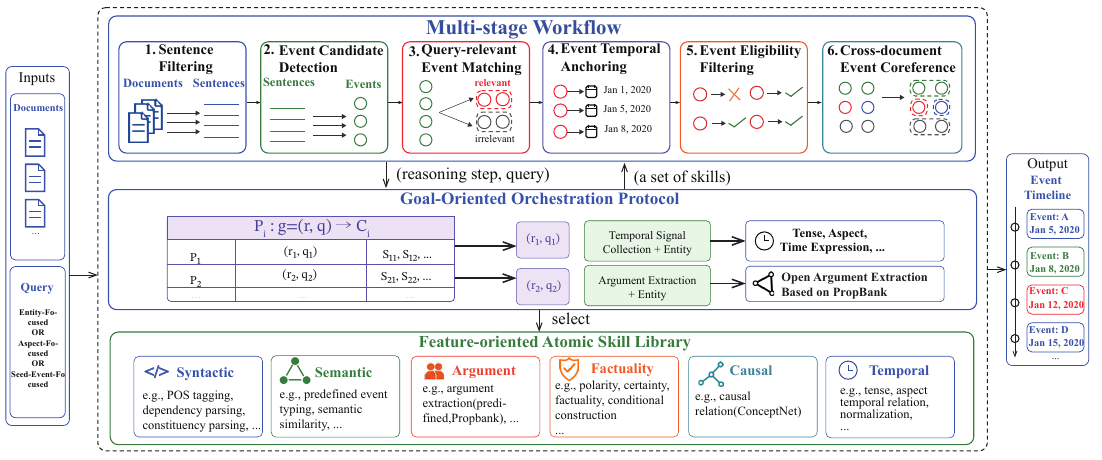}
    \caption{Our proposed TimelineAgent framework.}
    \label{fig:framework}
\end{figure*}

\section{Method}
As illustrated in Figure~\ref{fig:framework}, TimelineAgent decomposes QUEST into multiple sequential stages, where each agent performs stage-specific reasoning. During reasoning, LLM agents extract event features by invoking skills through the goal-oriented orchestration protocol. The protocol combines different reasoning sub-objectives and query settings into predefined goals, each associated with specific skill compositions and configurations, enabling flexible adaptation to diverse scenarios.

\subsection{Multi-stage Workflow}
\label{sec:method.1}
We design a unified workflow in which agents progressively leverage different event features and reasoning strategies to identify, filter, temporally anchor, and aggregate events. This staged design decouples complex reasoning processes while progressively refining intermediate event representations throughout timeline construction.

\paragraph{Sentence Filtering}
In this stage, the agent identifies potentially relevant sentences from the input document collection: \(\mathcal{S} = \{s_1, s_2, \ldots, s_m\}, \quad s_i \in D\).
The agent primarily relies on coarse-grained query-aware screening to preserve high-recall evidence for downstream timeline construction while reducing irrelevant information. For example, under entity-focused queries, the filtering process mainly retains sentences mentioning the target entity or its coreferential expressions, without requiring explicit event-level reasoning.

\paragraph{Event Candidate Detection}
Given the filtered sentence set $\mathcal{S}$, the agent identifies candidate event mentions $\mathcal{E} = \{e_1, e_2, ..., e_n\}$, which serve as a shared event space across different query focuses.
The agent primarily analyzes syntactic structures, including POS tags and constituency structures, and then applies predefined span-level constraints together with lexical knowledge (e.g., Collins English Dictionary) to maintain event boundary consistency. Finally, dependency parsing is applied to filter out light verbs and copula verbs.

\paragraph{Query-relevant Event Matching}
The agent leverages different forms of correspondence to distinguish query-relevant events from the candidate event set, resulting in \(\mathcal{E}_q \subseteq \mathcal{E}\).

For entity-focused queries, the agent first performs open-domain argument extraction based on semantic frame knowledge (e.g., PropBank~\cite{palmer-etal-2005-proposition}), and then models associations between query entities and event participants through entity coreference and bridging anaphora.

For aspect-focused queries, the agent first classifies events based on predefined aspect-oriented semantic types, and then identifies causally preceding events based on external causal knowledge (e.g., ConceptNet~\cite{10.5555/3298023.3298212}). Incorporating causal dependencies enables the timeline to provide more informative narratives, as illustrated by the aspect-focused example in Figure~\ref{fig:timeline}, which links the initial earthquake to downstream human-impact events such as deaths and injuries.

For seed-event-focused queries, the agent first determines event semantic types and then identifies related events through commonsense reasoning and external causal knowledge (e.g., ConceptNet). It further extracts arguments using predefined argument schemas and external frame knowledge (e.g., FrameNet), and evaluates argument consistency based on semantic roles.

\paragraph{Event Temporal Anchoring}
The agent grounds query-relevant events into a temporally anchored event set \(\mathcal{E}_t = \{(e_i, \tau_i)\}\), where \(\tau_i\) denotes the normalized timestamp of event \(e_i\).
The agent first collects diverse temporal signals, including tense, aspect, temporal expressions, and ordering cues, and then performs temporal inference to determine event timestamps through date calculation and temporal normalization. Document creation time (DCT) serves as a critical temporal reference point for integrating fragmented information across multiple documents~\cite{wang-etal-2022-dct}. During inference, reasoning over the temporal relation between events and DCT plays an important role, with different decision strategies required under different query types.

\paragraph{Event Eligibility Filtering}
This stage determines whether a query-relevant event should be included in the final timeline. It takes the temporally anchored event set \(\mathcal{E}_t\) as input and produces the eligible event set \(\mathcal{E}_{\text{eligible}} \subseteq \mathcal{E}_t\).\footnote{This stage is performed after temporal anchoring, since some queries depend on explicit temporal ranges.}
Event eligibility is separated from query relevance because it depends not on specific query content, but on query types and general timeline requirements. These include factuality-related dimensions, such as polarity, certainty, and factual status, as well as temporality constraints, which are typically reflected through semantic features. In addition, some query types impose explicit temporal range constraints on eligible events.

\paragraph{Cross-document Event Coreference Resolution}
Given \(\mathcal{E}_{\text{eligible}}\), the agent groups cross-document event mentions into clusters \(\mathcal{C}\), where events in the same cluster refer to the same real-world event.
The agent first determines semantic type compatibility between events, and then further evaluates argument consistency~\cite{barhom-etal-2019-revisiting}. Basic event arguments are extracted through semantic frame knowledge, such as PropBank and FrameNet, while previously inferred timestamps are also incorporated for temporal compatibility assessment.

An illustrative example of the workflow is provided in Appendix~\ref{sec:app_example}.

\subsection{Feature-oriented Skill Library}
We design an LLM-based feature-oriented skill library to support more comprehensive and robust event modeling across diverse reasoning stages and query settings. We decompose heterogeneous event features into reusable atomic skills organized by event attributes, while allowing flexible configurations such as external knowledge sources.

\paragraph{Syntactic}
In addition to basic syntactic analysis capabilities, such as part-of-speech tagging, constituency parsing, and dependency parsing~\cite{yan-etal-2019-event,lai-etal-2020-event,wan2024dependency}, we further design a skill based on lexical knowledge (e.g., Collins English Dictionary) to determine the extent of phrase-level constructions.

\paragraph{Semantic}
Semantic skills mainly support two settings: predefined event semantic type classification~\cite{doddington-etal-2004-automatic,wang-etal-2020-maven} and open-domain semantic similarity modeling between events~\cite{gao-etal-2022-improving}.

\paragraph{Argument}
Argument skills mainly support argument extraction under different schemas, including predefined schemas~\cite{ren-etal-2023-retrieve,shuang-etal-2024-thinking}, FrameNet-based~\cite{parekh-etal-2023-geneva}, and PropBank-based schemas~\cite{zhang-etal-2022-transfer}. They also provide alignment mechanisms between arguments or entities, including entity coreference resolution~\cite{khosla-etal-2021-evaluating}, and bridging anaphora~\cite{hou-2020-bridging}. The first two are formulated as binary matching tasks, while bridging anaphora is modeled as a multi-class relation classification task covering indirect relations, such as part--whole, set--member, and functional or attributive relations.

\paragraph{Factuality}
Factuality skills include event polarity, certainty, factuality, and conditionality, all formulated as binary classification tasks~\cite{Saur2009FactBankAC,sauri-pustejovsky-2012-sure,lee-etal-2015-event,rovera-etal-2025-modafact}. Among them, polarity is relatively objective, while the latter categories are more subjective. These skills can be flexibly combined to support more complex factuality judgments that vary across query settings. For example, in entity-focused queries, future events modified by \textit{will} are treated as factual events for timeline construction, whereas events expressed in conditional constructions are marked as conditional and excluded from the timeline.

\paragraph{Temporal}
Temporal skills are divided into signal-oriented and reasoning-oriented subcategories. Signal-oriented skills include temporal expressions, tense, aspect, and ordering cues~\cite{pustejovsky-etal-2010-iso,derczynski-gaizauskas-2015-temporal}. Reasoning-oriented skills include temporal normalization, date calculation, and temporal relation classification~\cite{wang-etal-2022-dct,xiong-etal-2024-large}.

\paragraph{Causal}
Causal category consists of causal relation classification based on causal knowledge from ConceptNet~\cite{10.5555/3298023.3298212,su-etal-2025-enhancing}.

\subsection{Goal-oriented Orchestration Protocol}
Instead of exposing the entire skill library to agents, which may introduce skill-level attention distraction and context pollution, we design an explicit orchestration protocol for controlled skill routing. The protocol is formalized as a structured coordination mechanism that maps reasoning goals to predefined skill compositions and configurations.

Formally, a reasoning goal is defined as:
$
g = (r, q)
$
where $r$ denotes a reasoning substep and $q$ denotes the query type. A protocol entry defines a mapping:
$
P_i : g \rightarrow C_i,
$
where $C_i$ denotes the orchestration configuration for the target reasoning goal.
Each configuration is represented as:
$
C_i = (O, \Phi),
$
where $O$ specifies the skill composition, such as combining tense, aspect, and temporal expressions for temporal reasoning, and $\Phi$ denotes skill-level configurations, such as selecting PropBank or FrameNet as the argument schema.

Detailed implementation of TimelineAgent is provided in Appendix~\ref{sec:appendix2.1}.

\section{Experiments}
We construct a benchmark covering three query settings for evaluating QUEST: entity, aspect, and seed-event queries. We compare against ICL and SFT baselines and a standard multi-agent framework, and further analyze the effects of each component and of staged reasoning.

\subsection{Experimental Setup}

\paragraph{Benchmark Construction}
Existing public datasets for query-centric event timeline extraction are very limited.\footnote{The dataset in ChemoTimelines 2024~\cite{yao-etal-2024-overview} task cannot be publicly released due to Data Use Agreements.} Due to the high annotation complexity of timeline construction, we construct our benchmark based on existing event datasets.

For entity-focused queries, we use SemEval-2015 Task 4~\cite{minard-etal-2015-semeval} dataset, which contains manually annotated entity-based event timelines from news articles. Entities include persons, organizations, products and financial entities.

For aspect-focused queries, we construct a seven-category crisis-oriented aspect schema based on humanitarian information categories from social media scenarios~\cite{imran2016twitter,vieweg2010microblogging, castillo2011credibility}. We extend the EventStoryLine Corpus~\cite{caselli-vossen-2017-event}, which models the development of news stories at the document level. We select crisis-related topics, including earthquakes, prisoner escapes, bank blast fires, and cable outages, manually classify events into aspect types, and transform existing temporal and causal relations into document-level timelines. Finally, we incorporate cross-document event coreference annotations from ECB+~\cite{cybulska-vossen-2014-using} to construct cross-document timelines.  The event classification process was conducted by two linguistics students, achieving a Cohen's kappa coefficient of 0.85. Details of the aspect types and the document-level timeline transformation process are provided in Appendix~\ref{sec:appb.1}.

For seed-event-focused queries, we further extend PledgeTracker~\cite{chen-etal-2025-pledgetracker}, which annotates political pledges and their associated timelines for pledge fulfillment tracking. While their task setting is closely related to ours, timeline events are represented as sentence-level summarized descriptions. We instead use political pledges as seed events and further annotate action-level event mentions grounded in the original source documents based on their event descriptions.
The action-level event annotation is based on SemEval-2015 Task 4 annotation guidelines.\footnote{\href{https://alt.qcri.org/semeval2015/task4/data/uploads/documentation/manualannotationguidelines-task4.pdf}{https://alt.qcri.org/semeval2015/task4/data/uploads/\\documentation/manualannotationguidelines-task4.pdf}} Two linguistics students participated in the annotation, achieving 81\% span-level F1 agreement.

\begin{table}[t]
\centering
\resizebox{0.9\columnwidth}{!}{%
\begin{tabular}{lccc}
\hline
                        & Entity & Aspect$^\dagger$ & Seed-event$^\dagger$ \\ \hline
\# documents            & 120    & 46     &  1475          \\
\# events               & 1102   & 647    &  406          \\
\# timelines            & 42     & 28     &  26          \\
events / timeline       & 26.3   & 23.1   &  15.6          \\
event chains / timeline & 19.9   & 13.9       &   11.1         \\
docs / timeline         & 6.9    & 8.6    &  6.0          \\ \hline
\end{tabular}%
}
\caption{Dataset statistics of our benchmark. $^\dagger$ denotes datasets newly annotated in this work.}
\label{tab:dataset_stats}
\end{table}

Dataset statistics are summarized in Table~\ref{tab:dataset_stats}. Illustrative annotation examples for the seed-event- and aspect-focused query settings are provided in Appendix~\ref{sec:annotation_examples}.

\paragraph{Evaluation Metric}

We follow the official SemEval-2015 TimeLine evaluation protocol~\cite{minard-etal-2015-semeval}. Rather than scoring isolated temporal decisions, the protocol transforms each predicted and gold timeline into a temporal graph whose nodes are event mentions and normalized time anchors and whose edges encode event--time anchoring and event--event relations. Graph-based precision, recall, and F1 jointly evaluate event selection, temporal grounding, and cross-document event coreference decisions. Further details of the metric are provided in Appendix~\ref{sec:appb.2}.

\paragraph{Baseline Methods}
We compare TimelineAgent with a traditional hand-crafted pipeline system (DLT~\cite{laparra-etal-2015-document}), end-to-end LLM settings using in-context learning (ICL) and supervised fine-tuning (SFT), and the task-agnostic multi-agent framework TDAG~\cite{Wang2024TDAGAM}. TDAG is a strong task-agnostic multi-agent framework that has demonstrated superior performance over existing agent paradigms such as ReAct~\cite{yao2022react} and Plan-and-Solve~\cite{wang-etal-2023-plan} on complex reasoning tasks. Detailed baseline implementations are provided in Appendix~\ref{sec:a.2}.

\paragraph{Implementation Details}
We employ two backbone LLMs in our experiments: DeepSeek-V3~\cite{deepseekai2025deepseekv3technicalreport} and Qwen3-32B~\cite{Yang2025Qwen3TR}. DeepSeek-V3 is a 671B-parameter language model accessed through its API for inference. Qwen3-32B is deployed on 4 NVIDIA A100 GPUs with 80GB memory for both fine-tuning and inference. To reduce randomness and ensure stable performance, we set the decoding temperature to 0. For SFT, we use the original trial set (6 timeline instances) for entity-focused queries. To maintain a comparable training scale, we construct five data splits for aspect-focused and seed-event-focused queries, with 6 and 5 timeline instances used for training in each split, respectively, while the remaining instances are used for test. Detailed configurations are provided in Appendix~\ref{sec:appb.3}.

\subsection{Main Results}

\begin{table}[t]
\centering
\resizebox{\columnwidth}{!}{%
\begin{tabular}{lccccccccc}
\hline
\multirow{2}{*}{Method} & \multicolumn{3}{c}{Entity} & \multicolumn{3}{c}{Aspect} & \multicolumn{3}{c}{Seed-event} \\ \cline{2-10}
              & P    & R    & F1            & P    & R    & F1            & P    & R    & F1            \\ \hline
DLT           & 20.7 & 11.0 & 14.3          & -    & -    & -             & -    & -    & -             \\ \hline
\multicolumn{10}{l}{\textit{Qwen3-32B}}                                                                 \\
ICL           & 17.8 & 12.1 & 14.4          & 28.1 & 23.8 & 25.8          & 8.3  & 12.9 & 10.1          \\
TDAG          & 15.3 & 20.8  & 17.6          & 33.5 & 27.5 & 30.2          &  11.3 & 19.6 & 14.3          \\
SFT           & 24.0 & 27.9 & 25.8          & 40.2 & 31.8 & 35.5          & 17.6 & 25.8 & 20.9          \\
TimelineAgent & 29.2 & 35.7 & \underline{32.1} & 41.3 & 36.8 & \underline{38.9} & 22.9 & 33.8 & \underline{27.3} \\ \hline
\multicolumn{10}{l}{\textit{DeepSeek-V3}}                                                               \\
ICL           & 21.2 & 15.0 & 17.5          & 33.7 & 24.2 & 28.2          & 9.8  & 17.4 & 12.5          \\
TDAG          & 22.2 & 44.3 & 29.6          & 40.3 & 33.2 & 36.4          & 19.5 & 30.2 & 23.7          \\
TimelineAgent & 37.2 & 38.6 & \textbf{37.9} & 45.8 & 40.3 & \textbf{42.9} & 27.8 & 40.9 & \textbf{33.1} \\ \hline
\end{tabular}%
}
\caption{Performance comparison on QUEST.}
\label{tab:main_results}
\end{table}

The overall performance comparison is shown in Table~\ref{tab:main_results}.
TimelineAgent consistently outperforms all baselines across different query settings and backbone model scales, demonstrating the effectiveness of our framework. In particular, our Qwen3-32B model surpasses the general-purpose agent framework TDAG built on DeepSeek-V3, indicating the importance of task-specific reasoning organization and event modeling for QUEST. However, overall performance remains around 30--40\% F1, highlighting the inherent difficulty of QUEST.

Among different query settings, seed-event-focused queries are the most challenging, followed by entity-focused queries, while aspect-focused queries are relatively easier. This trend is roughly proportional to the sparsity of relevant events in documents, with an average of 0.3, 9.2, and 14.1 relevant events per document, respectively, highlighting the inherent challenges of cross-document event  reasoning in QUEST.

Compared with ICL, SFT substantially improves performance, indicating the benefit of task-specific adaptation. However, its effectiveness is still limited by the low-resource training data. TDAG, as a general-purpose multi-agent framework, further demonstrates the effectiveness of task decomposition and coordinated agent reasoning. Nevertheless, both approaches still show a considerable gap from TimelineAgent, highlighting the effectiveness of our task-specific understanding and structured modeling of the complex requirements in QUEST. Larger backbone models consistently outperform smaller ones across all settings, suggesting the importance of strong general reasoning capabilities.

\subsection{Ablation Study}
We first conduct architectural ablations by removing different framework components to analyze their contributions. We further perform stage-wise ablations by replacing specialized reasoning with simple single-step prompting. In this setting, each agent is not equipped with explicit skills or multi-step reasoning processes. Finally, we conduct error propagation analysis by replacing stage outputs with gold annotations to investigate the downstream impact of errors.

\begin{figure}[t]
    \centering
    \includegraphics[width=1\linewidth]{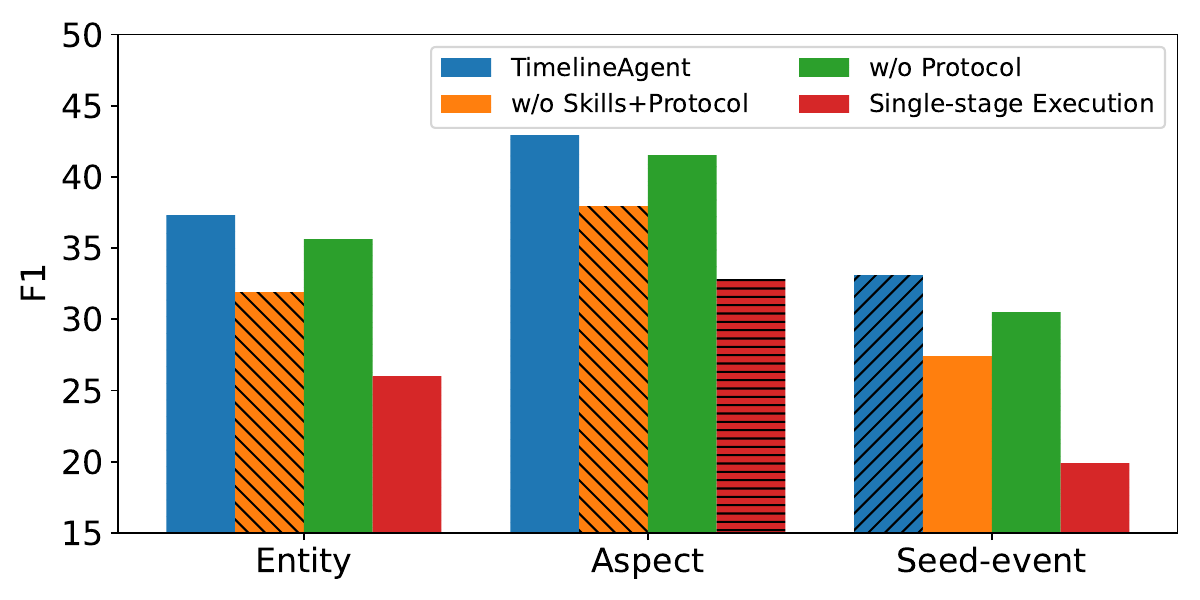}
    \caption{Effects of skill library,  orchestration protocol and staged execution of TimelineAgent.}
    \label{fig:arc_ablation}
\end{figure}

\paragraph{Architectural Ablation.}
As shown in Figure~\ref{fig:arc_ablation}, removing both skills and orchestration protocols (w/o Skills+Protocol) leads to substantial performance degradation across all query settings, demonstrating that merely providing high-level reasoning objectives without specialized event feature modeling is insufficient for event-centric reasoning. When only orchestration protocol is removed (w/o Protocol), the model can still utilize useful skills to some extent, but performance remains consistently inferior to the full model, indicating that directly exposing all skills as flat context is less effective than our structured skill routing strategy.
We observe that preserving both skills and protocols while executing the entire workflow in a single unified stage (Single-stage Execution) results in particularly severe performance degradation. This finding highlights the importance of staged execution for handling the overall reasoning complexity.

\begin{figure}[t]
    \centering
    \includegraphics[width=1\linewidth]{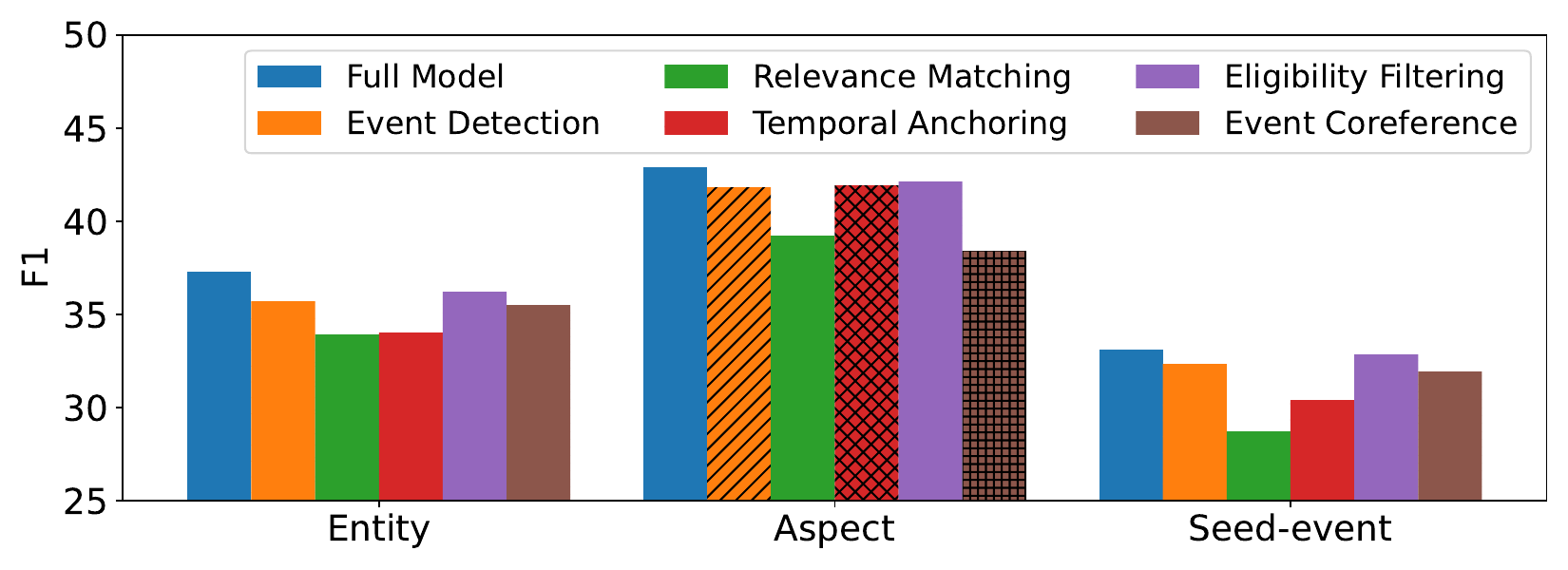}
    \caption{Effects of different workflow stages of TimelineAgent.}
    \label{fig:stage_ablation}
\end{figure}

\paragraph{Stage-wise Ablation.}
As shown in Figure~\ref{fig:stage_ablation}, event detection consistently contributes to all query settings, demonstrating the effectiveness of our fine-grained syntactic modeling. Relevance matching yields the largest overall contribution across all settings, highlighting the importance of our structured reasoning and decomposed feature modeling. Temporal anchoring is particularly important for entity-focused and seed-event-focused queries, reflecting the challenges of reasoning over long temporal spans and temporally evolving events. Eligibility filtering also provides consistent improvements, indicating that modeling factuality and semantic temporality is beneficial for identifying temporally eligible events. Event coreference contributes most substantially to aspect-focused queries, demonstrating the effectiveness of our modeling for distinguishing highly similar events occurring within relatively compact temporal contexts.

\paragraph{Error Propagation Analysis.}
Since the datasets do not separately provide annotations for general events, event relevance, and event eligibility, we combine them into a unified stage referred to as \emph{event detection, matching, and filtering}. As shown in Table~\ref{tab:error_prop}, event detection, matching, and filtering constitute the most critical stage across all query settings, indicating that accurately identifying and selecting relevant events is fundamental for timeline construction. Temporal anchoring is particularly important for entity-focused and seed-event-focused queries, which typically involve broader temporal spans and require more complex temporal reasoning. In contrast, aspect-focused queries are more sensitive to event coreference, since they are centered around different aspects of a relatively compact seminal event with many repeated event mentions across documents. We provide additional qualitative error analysis in Appendix~\ref{sec:app3.3}.

We provide case studies of key stages in Appendix~\ref{sec:app3.2}.

\begin{table}[t]
\centering
\resizebox{\columnwidth}{!}{%
\begin{tabular}{lccc}
\hline
\textbf{Oracle Setting}                & \textbf{Entity} & \textbf{Aspect} & \textbf{Seed-event} \\ \hline
TimelineAgent                           & 37.9            & 42.9            & 33.1                \\
Gold Detection, Matching, \& Filtering & 67.6            & 69.8            & 71.4                \\
Gold Temporal Anchoring                & 55.6            & 52.4            & 41.0                \\
Gold Event Coreference                 & 42.1            & 56.1            & 39.4                \\ \hline
\end{tabular}%
}
\caption{Error propagation analysis using gold annotations for different workflow stages.}
\label{tab:error_prop}
\end{table}

\subsection{Impact of LLM Backbones}
\begin{figure}[t]
    \centering
    \includegraphics[width=1\linewidth]{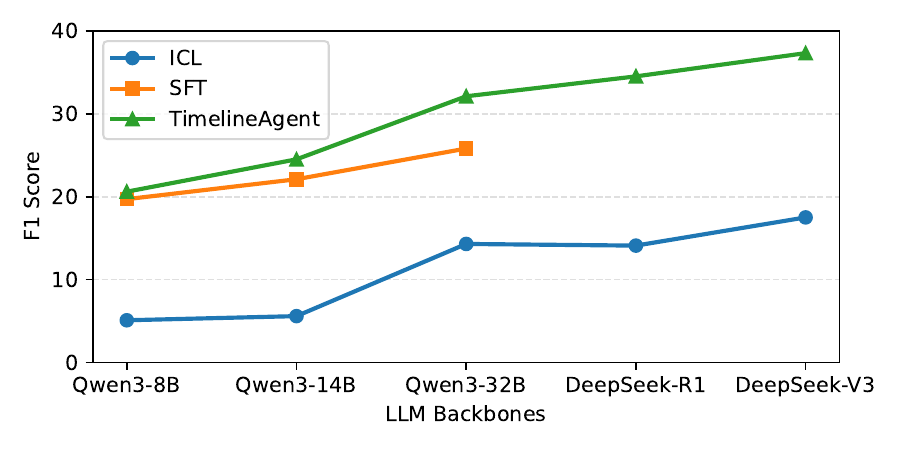}
    \caption{Comparison of performance across different LLM backbones under entity setting.}
    \label{fig:llm_comparison}
\end{figure}

We further compare different backbone model scales to analyze the relationship between model capacity and performance, as shown in Figure~\ref{fig:llm_comparison}. Overall, larger models consistently achieve better results. However, the performance gains gradually diminish when scaling from Qwen3-32B to DeepSeek-V3 (671B), suggesting that model scale alone is no longer the dominant factor for QUEST, while task-specific reasoning and modeling become increasingly important.
Compared with SFT, our framework achieves slightly better performance on smaller models such as 8B and 14B, but demonstrates larger gains as model size further increases to 32B. This trend suggests that stronger LLM reasoning capabilities are more effective for handling decomposed reasoning subproblems in our framework, while directly fine-tuning models to perform complex end-to-end reasoning remains difficult under extremely low-resource conditions with only 5--6 training instances. Moreover, unlike SFT, our framework does not require training separate task-specific models for different query settings, enabling stronger generalization across diverse timeline construction scenarios.

\subsection{Impact of SFT Training Scale}
\label{sec:sft_scale}
To assess whether the low-resource setting of the SFT baseline accounts for its performance gap, we increase the amount of supervision at medium and large scales. Table~\ref{tab:sft_scale} shows that increasing supervision consistently improves SFT performance across all three query settings, although the gains diminish as the training set grows. At the large scale, SFT remains below Qwen3-32B-based TimelineAgent on Entity and Seed-event queries, but exceeds it by 0.6 F1 on Aspect queries. TimelineAgent with DeepSeek-V3 outperforms large-scale SFT in every setting.  Thus, while end-to-end SFT benefits from additional supervision, TimelineAgent remains competitive in limited-supervision settings and can benefit from larger foundation models without task-specific fine-tuning.

\begin{table}[t]
\centering
\small
\resizebox{\columnwidth}{!}{%
\begin{tabular}{llccc}
\hline
Backbone & Method & Entity & Aspect & Seed-event \\ \hline
\multirow{5}{*}{Qwen3-32B}
& ICL           & 14.4 & 25.8 & 10.1 \\
& SFT           & 25.8 & 35.5 & 20.9 \\
& SFT (medium)  & 28.6 & 38.0 & 24.4 \\
& SFT (large)   & 29.9 & 39.5 & 26.2 \\
& TimelineAgent & 32.1 & 38.9 & 27.3 \\ \hline
DeepSeek-V3 & TimelineAgent & 37.9 & 42.9 & 33.1 \\ \hline
\end{tabular}%
}
\caption{F1 scores at different supervised training scales for end-to-end SFT. The standard, medium, and large SFT settings use 6/6/5, 18/12/10, and 30/18/15 training timelines for Entity, Aspect, and Seed-event queries, respectively.}
\label{tab:sft_scale}
\end{table}

\subsection{Computational Efficiency}
To improve computational efficiency, we first perform sentence filtering to remove irrelevant context before event reasoning. This step filters out approximately 50\%, 30\%, and 90\% of sentences for three query settings, respectively.
To further reduce inference overhead, we jointly process all events within the same sentence at each step rather than reasoning over them individually. Compared with standard multi-stage ICL approaches that typically execute a single reasoning step per stage, our framework requires only
$
\frac{2 + \frac{2+2+3}{3} + 2 + 2 + 2}{5}
\approx 2.1\times
$
more reasoning executions on average. Compared with the automatic multi-agent decomposition framework TDAG, which incurs approximately 4$\times$ more executions, our framework remains more efficient.
In practice, a full inference run takes around ten hours depending on the batching configuration, demonstrating the practical feasibility of our framework.

\section{Conclusion}
We introduced the practical yet challenging task of QUEST, which requires handling complex processing pipelines and diverse event features under different query settings. We propose TimelineAgent, a multi-agent framework that decomposes timeline construction into coordinated reasoning stages supported by reusable atomic skills and goal-oriented orchestration protocols. Experimental results across multiple query settings demonstrate the effectiveness of our framework over strong baselines. We hope our framework can provide a practical direction for applying structured agent systems to other complex information extraction tasks.

\section*{Limitations}

Our work mainly focuses on establishing a principled framework for the complex QUEST task. Based on our atomic feature-oriented skill design and goal-oriented orchestration principles, future work could explore finer-grained orchestration strategies, such as independently invoking atomic skills for individual event mentions to reduce the complexity of each reasoning step and enable more precise reasoning and feature modeling. To address the substantially higher computational overhead introduced by such designs, future research may investigate hybrid architectures that combine large and small language models, where stronger models handle more demanding reasoning processes while lightweight models perform relatively simpler feature extraction tasks such as syntactic analysis and semantic categorization. Such designs may provide stronger reasoning capabilities while maintaining better computational efficiency.

Constructing high-quality structured event timelines is an annotation-intensive process. To scale the benchmark while maintaining annotation quality, human-agent collaboration offers a practical direction. Under different query requirements, human annotators can edit the reasoning logic for timeline construction and customize the corresponding skills and orchestration protocols, enabling more flexible extension of the benchmark. Meanwhile, the agent's interpretable reasoning process allows annotators to provide supervision and corrections at intermediate steps, helping ensure annotation quality while scaling the benchmark.

\section*{Acknowledgments}
We would like to thank anonymous reviewers for their insightful comments to help improve the paper.
This work has been supported by the National Natural Science Foundation of China (NSFC) Key Project under Grant Number 62336006.
Andreas Vlachos and Yulong Chen were supported by the DARPA program SciFy.

\bibliography{custom}

\appendix

\section{Method Details}
\label{sec:appendix2}

\subsection{Example of Multi-stage Timeline Construction}
\label{sec:app_example}

Consider an entity-focused query centered on \textit{Steve Jobs}.
The workflow first performs query-relevant sentence filtering to identify potentially useful context from the document collection. Sentences mentioning the target entity or its coreferential expressions are retained as high-recall evidence for downstream reasoning.

The agent then performs query-relevant event matching to identify events associated with the query entity. Candidate events such as \textit{announced}, \textit{fought}, \textit{resigned}, and \textit{unveiled} are extracted from the filtered sentences. Using semantic frame knowledge and entity coreference reasoning, the agent determines whether event participants are associated with the target entity. For example, the event \textit{unveiled} is retained because the pronoun ``he'' in its argument structure is resolved to ``Steve Jobs.''

Next, the agent performs event temporal anchoring. Temporal signals including tense, aspect, temporal expressions, and temporal ordering cues are first extracted from context. For example, given the sentence ``Jobs announced the product on Monday'' with document creation time \texttt{2011-01-17}, the agent extracts the temporal expression ``Monday'' and normalizes the event timestamp to \texttt{2011-01-17}. Similarly, the sentence ``Jobs unveiled the device in 2004'' leads to the normalized timestamp \texttt{2004-XX-XX} through temporal expression normalization.

After temporal grounding, the agent performs cross-document event coreference resolution. Event mentions are first compared based on semantic type compatibility and semantic similarity. The agent then evaluates argument consistency and temporal compatibility using extracted participants and normalized timestamps. For example, two event mentions containing the trigger \textit{announced} may initially appear semantically compatible. However, if one event corresponds to ``Steve Jobs announced Lion OS X'' on \texttt{2010-10-23}, while another corresponds to ``Steve Jobs announced medical leave'' on \texttt{2011-01-17}, the agent determines that the arguments and timestamps are inconsistent and therefore assigns them to different event clusters.

Finally, the clustered and temporally grounded events are aggregated into a chronologically ordered timeline. Through this staged workflow, TimelineAgent progressively transforms noisy cross-document evidence into structured event timelines with explicit temporal organization.

\subsection{Implementation Details of TimelineAgent}
\label{sec:appendix2.1}

\subsubsection{Multi-stage Workflow}

\begin{tcolorbox}[
title=Sentence Filtering Stage,
colback=gray!5,
colframe=gray!60,
fonttitle=\bfseries,
breakable]

\textbf{Objective:}

Perform coarse-grained query-aware evidence screening to preserve
high-recall candidate evidence for downstream timeline construction.

\vspace{0.3em}
\textbf{Input:}

\begin{verbatim}
Query:
{query}

Document:
{document}
\end{verbatim}

\vspace{0.3em}
\textbf{Query-specific Focus:}
\begin{itemize}
    \item Entity-focused: entity mentions and aliases
    \item Aspect-focused: semantic topical consistency
    \item Seed-event-focused: triggered by the seed event
\end{itemize}

\vspace{0.3em}
\textbf{Output:}

Filtered sentence set
$
\mathcal{S}' \subseteq \mathcal{S}.
$

\end{tcolorbox}

\begin{tcolorbox}[
title=Event Candidate Detection Stage,
colback=gray!5,
colframe=gray!60,
fonttitle=\bfseries,
breakable]

\textbf{Objective:}

Identify structurally valid event candidates from filtered sentences while preserving high event recall for downstream reasoning.

\vspace{0.5em}
\textbf{Input:}

Sentence set
$
\mathcal{S}' = \{s_1, s_2, \ldots, s_m\}
$
from the sentence filtering stage.

\vspace{0.5em}
\textbf{Step 1: Syntactic Event Analysis}

\textbf{Input Placeholder:}
\begin{verbatim}
Sentence:
{sentence}

\end{verbatim}

The first prompt performs coarse-grained syntactic analysis to identify potential event candidates based on part-of-speech categories, constituency structures, and dependency relations.

\begin{itemize}
    \item Identify candidate verbal, nominal, and predicative event mentions
    \item Analyze phrase structures and dependency heads
    \item Detect possible multi-token event constructions
\end{itemize}

\vspace{0.3em}
\textbf{Activated Skills:}
\begin{itemize}
    \item POS tagging
    \item Constituency parsing
    \item Dependency parsing
\end{itemize}

\vspace{0.5em}
\textbf{Intermediate Output:}

Candidate event set
$
\mathcal{E}' = \{e_1, e_2, \ldots, e_k\}
$
with syntactic spans and structural attributes.

\vspace{0.5em}
\textbf{Step 2: Rule-guided Event Filtering}

\textbf{Input Placeholder:}
\begin{verbatim}
Sentence:
{sentence}

Candidate Events:
{candidate_events}

Syntactic Analysis:
{syntactic_outputs}
\end{verbatim}

The second prompt applies rule-based filtering and lexical constraints to refine event candidates according to annotation-oriented syntactic criteria.

\begin{itemize}
    \item Remove non-eventive syntactic heads
    \item Filter auxiliary verbs, copular verbs, and light verb constructions
    \item Verify lexicalized multi-word expressions using Collins English Dictionary
    \item Resolve uncertain phrase-level event boundaries
\end{itemize}

\vspace{0.3em}
\textbf{Activated Skills:}
\begin{itemize}
    \item Lexical phrase validation
\end{itemize}

\vspace{0.5em}
\textbf{Output:}

Structured event candidate set:
$
\mathcal{E} = \{e_1, e_2, \ldots, e_n\}
$
with event spans and syntactic attributes.

\end{tcolorbox}

\begin{tcolorbox}[
title=Query-relevant Event Matching Stage,
colback=gray!5,
colframe=gray!60,
fonttitle=\bfseries,
breakable]

\textbf{Objective:}

Identify query-relevant events from the candidate event set
$
\mathcal{E}
$
through semantic, argument, and causal correspondence reasoning.

\vspace{0.3em}
\textbf{Input:}

Candidate event set
$
\mathcal{E} = \{e_1, e_2, \ldots, e_n\}.
$

\vspace{0.5em}
\textbf{Input Placeholder:}

\begin{verbatim}
Query:
{query}

Query Type:
{query_type}

Sentence:
{sentence}

Candidate Events:
{event_triggers}

Context:
{context}

Optional Knowledge Source:
{knowledge_source}
\end{verbatim}

\vspace{0.3em}
\textbf{General Reasoning Procedure:}
\begin{itemize}
    \item Event semantic correspondence reasoning
    \item Argument and participant consistency reasoning
    \item Commonsense and causal dependency reasoning
\end{itemize}

\vspace{0.3em}
\textbf{Output:}

Query-relevant event set
$
\mathcal{E}_q \subseteq \mathcal{E}.
$

\end{tcolorbox}

\paragraph{Entity-focused Query} \

\textbf{Goal:}
Identify events semantically or argumentatively associated with the target entity.

\textbf{Reasoning Procedure:}
\begin{itemize}
    \item Extract event arguments based on open-domain semantic frames
    \item Identify associations between query entities and event participants. Resolve entity coreference and bridging relations
\end{itemize}

\textbf{Skills:}
\begin{itemize}
    \item Argument extraction (PropBank)
    \item Entity coreference resolution
    \item Bridging anaphora reasoning
\end{itemize}

\paragraph{Aspect-focused Query}\

\textbf{Goal:}
Identify events relevant to predefined crisis-oriented aspects.

\textbf{Reasoning Procedure:}
\begin{itemize}
    \item Classify events into predefined aspect semantic categories
    \item Identify causally related preceding events. Construct informative aspect-centered narratives
\end{itemize}

\textbf{Skills:}
\begin{itemize}
    \item Aspect semantic type classification
    \item Causal relation reasoning. External causal knowledge (ConceptNet)
\end{itemize}

\paragraph{Seed-event-focused Orchestration} \

\textbf{Goal:}
Identify events semantically and causally related to the seed event.

\textbf{Reasoning Procedure:}
\begin{itemize}
    \item Determine event semantic types. Identify commonsense and causal event relations
    \item Extract arguments
    \item Evaluate semantic role consistency
\end{itemize}

\textbf{Skills:}
\begin{itemize}
    \item Event semantic typing
    \item Commonsense and causal reasoning
    \item Argument extraction (FrameNet)
\end{itemize}

\begin{tcolorbox}[
title=Event Temporal Anchoring Stage,
colback=gray!5,
colframe=gray!60,
fonttitle=\bfseries,
breakable]

\textbf{Objective:}

Ground query-relevant events into normalized timestamps to construct the temporally anchored event set
$
\mathcal{E}_t = \{(e_i, \tau_i)\}.
$

\vspace{0.3em}
\textbf{Input:}

Query-relevant event set
$
\mathcal{E}_q.
$

\vspace{0.5em}
\textbf{Input Placeholder:}

\begin{verbatim}
Query:
{query}

Sentence:
{sentence}

Events:
{event_triggers}

Temporal Context:
{context}
\end{verbatim}

\vspace{0.3em}
\textbf{Step 1: Temporal Signal Extraction}
\begin{itemize}
    \item Extract tense and aspect information
    \item Identify temporal expressions and DCT
    \item Identify temporal ordering cues
\end{itemize}

\textbf{Skills:}
\begin{itemize}
    \item Tense analysis
    \item Aspect analysis
    \item Temporal expression extraction
    \item Temporal ordering cue extraction
\end{itemize}

\vspace{0.3em}
\textbf{Step 2: Temporal Inference}
\begin{itemize}
    \item Resolve relative temporal expressions
    \item Perform date calculation and temporal normalization
    \item Derive normalized timestamps under document-level temporal contexts
\end{itemize}

\textbf{Skills:}
\begin{itemize}
    \item Temporal relation classification
    \item Date calculation
    \item Temporal normalization
\end{itemize}

\vspace{0.3em}
\textbf{Output:}

Temporally anchored event set
$
\mathcal{E}_t = \{(e_i, \tau_i)\}.
$

\end{tcolorbox}

\begin{tcolorbox}[
title=Cross-document Event Coreference Resolution Stage,
colback=gray!5,
colframe=gray!60,
fonttitle=\bfseries,
breakable]

\textbf{Objective:}

Group eligible cross-document event mentions into event clusters
$
\mathcal{C},
$
where events within the same cluster refer to the same real-world event.

\vspace{0.3em}
\textbf{Input:}

Eligible event set
$
\mathcal{E}_{\text{eligible}}.
$

\vspace{0.5em}
\textbf{Input Placeholder:}

\begin{verbatim}
Sentence:
{sentence}

Event:
{event_trigger}

Event Type:
{event_type}

Arguments:
{arguments}

Timestamp:
{timestamp}

Context:
{context}
\end{verbatim}

\vspace{0.3em}
\textbf{Step 1: Semantic Compatibility Modeling}
\begin{itemize}
    \item Determine semantic type compatibility between events
    \item Evaluate semantic similarity between event mentions
    \item Filter semantically incompatible event pairs
\end{itemize}

\textbf{Skills:}
\begin{itemize}
    \item Semantic similarity reasoning
\end{itemize}

\vspace{0.3em}
\textbf{Step 2: Argument and Temporal Consistency Reasoning}
\begin{itemize}
    \item Extract event arguments through semantic frame knowledge
    \item Evaluate participant and semantic role consistency
    \item Incorporate temporal compatibility based on inferred timestamps
\end{itemize}

\textbf{Skills:}
\begin{itemize}
    \item Argument extraction (PropBank / FrameNet)
    \item Semantic role consistency modeling
    \item Temporal compatibility reasoning
\end{itemize}

\vspace{0.3em}
\textbf{Output:}

Cross-document event clusters
$
\mathcal{C}.
$

\end{tcolorbox}

\subsubsection{Feature-oriented Atomic Skills}
\label{appendix:prompt_examples}

Table~\ref{tab:skill_impl} shows skill formulations and configurations. All atomic skills are implemented as constrained prompting modules with structured inputs, configurations, and standardized outputs. To improve controllability and reduce reasoning interference, each skill operates under restricted objectives and predefined output schemas.

\begin{table*}[t]
\centering
\small
\begin{tabular}{p{2cm}p{7.2cm}p{5cm}}
\hline
\textbf{Category} & \textbf{Atomic Skills} & \textbf{Configurations} \\
\hline

Syntactic &
Part-of-speech tagging, constituency parsing, dependency parsing, phrase span detection &
Collins English Dictionary for phrase span detection \\

Semantic &
Event type classification, semantic similarity modeling &
predefined semantic categories for classification\\

Argument &
Argument extraction, entity coreference resolution, bridging anaphora &
PropBank schemas, FrameNet schemas, predefined schemas for artument extraction\\

Factuality &
Polarity classification, certainty classification, factuality classification, conditionality classification &
labels for certainty and factuality classification \\

Temporal &
Temporal expression extraction, tense analysis, aspect analysis, ordering cue extraction, temporal normalization, date calculation, temporal relation classification &
Temporal relation label sets for classification \\

Causal &
Causal relation classification &
ConceptNet causal knowledge \\

\hline
\end{tabular}
\caption{Atomic skills and configurations in the feature-oriented skill library.}
\label{tab:skill_impl}
\end{table*}

All skills return structured intermediate outputs in unified JSON-style formats, enabling flexible composition across workflow stages. The following table presents a representative prompt template used for atomic skills. The orchestration protocol dynamically injects task objectives, configuration constraints, and external knowledge specifications according to different reasoning goals.

\begin{tcolorbox}[
title=Atomic Skill Prompt Template,
colback=gray!5,
colframe=gray!60,
fonttitle=\bfseries,
breakable]
\textbf{System Instruction:}

You are an atomic event modeling skill expert.

\vspace{0.3em}
\textbf{Task Objective:}

\texttt{\{Task\_description\}}

\vspace{0.3em}
\textbf{Input Event:}

\texttt{\{event\_context\}}

\vspace{0.3em}
\textbf{Configuration:}

\texttt{\{Skill\_specific\_configuration\}}

\vspace{0.3em}
\textbf{Constraints:}
\begin{itemize}
    \item Follow the predefined label space.
    \item Only perform the assigned task.
    \item Return structured JSON outputs.
\end{itemize}

\vspace{0.3em}
\textbf{Output Format:}

\begin{verbatim}
{
  "event": "...",
  "prediction": "...",
  "confidence": "..."
}
\end{verbatim}
\end{tcolorbox}

Different skills instantiate this template with task-specific objectives and configurations. For example, temporal skills inject temporal normalization rules and relation labels, while argument extraction skills configure different semantic role schemas such as PropBank or FrameNet.

\subsubsection{Goal-oriented Orchestration Protocol}

The orchestration protocol selects and configures atomic skills according to reasoning substeps and query requirements. In practice, each protocol entry specifies two components: (1) the target reasoning objective and (2) the required skill composition.
The protocol configurations are shown in Table~\ref{tab:skills}, which illustrates how different workflow stages activate heterogeneous skills under different query settings.

\begin{table*}[t]
\centering
\resizebox{\textwidth}{!}{%
\begin{tabular}{ll}
\hline
\textbf{Pipeline Stage} &
  \textbf{Skills} \\ \hline
\begin{tabular}[c]{@{}l@{}}Event Candidate \\ Detection\end{tabular} &
  Syntactic (POS tagging, constituency parsing, dependency parsing, lexical resource (Collins English Dictionary)) \\ \hline
\begin{tabular}[c]{@{}l@{}}Query-relevant \\ Event Matching\end{tabular} &
  \begin{tabular}[c]{@{}l@{}}Semantic (predefined event semantic typing$^{2}$, semantic similarity$^{3}$) \\ Argument (argument extraction (predefined$^{3}$, PropBank$^{1}$, FrameNet$^{3}$), entity coreference resolution$^{1}$, bridging anaphora$^{1}$)\\ Causal (causal relation (ConceptNet$^{2,3}$))\end{tabular} \\ \hline
\begin{tabular}[c]{@{}l@{}}Event Temporal \\ Anchoring\end{tabular} &
  Temporal (tense, aspect, time expression, ordering cues, date calculation, temporal normalization, temporal relation) \\ \hline
\begin{tabular}[c]{@{}l@{}}Event Eligibility \\ Filtering\end{tabular} &
  \begin{tabular}[c]{@{}l@{}}Factuality (polarity, factuality, certainty, condition construction)\\ Temporal (date calculation$^{3}$)\\ Semantic (predefined semantic typing)\end{tabular} \\ \hline
\begin{tabular}[c]{@{}l@{}}Cross-document \\ Event Coreference \\ Resolution\end{tabular} &
  \begin{tabular}[c]{@{}l@{}}Semantic (semantic similarity)\\ Argument extraction (PropBank)\end{tabular} \\ \hline
\end{tabular}%
}
\caption{Goal-oriented orchestration configurations across workflow stages and query settings: Entity$^{1}$, Aspect$^{2}$, and Seed-event$^{3}$.}
\label{tab:skills}
\end{table*}

\subsection{Implementation Details for Baselines}
\label{sec:a.2}

\subsubsection{Implementation Details for End-to-End SFT and ICL}

\begin{tcolorbox}[title=Query-focused Event Timeline Construction Template, breakable]

\textbf{Task Objective}

Given a document collection and a query, construct a query-focused event timeline by identifying, temporally anchoring, ordering, and aggregating relevant events across documents.

\medskip

\textbf{Input Placeholder}
\begin{verbatim}
Documents:
{documents}

Query:
{query}

Query Type:
{query_type}
\end{verbatim}

\medskip

\textbf{Timeline Output Format}
\begin{verbatim}
ordering    time_anchor
docid-sentid-event
0           YYYY-MM-DD
docid-sentid-event
1           YYYY-MM-DD
docid-sentid-event
...
\end{verbatim}

\medskip

\hrule
\medskip

\textbf{Core Requirements}

\begin{itemize}
    \item Detect query-relevant events
    \item Resolve cross-document event coreference
    \item Infer normalized timestamps
    \item Construct temporally ordered timelines
\end{itemize}

\medskip

\textbf{Temporal Constraints}

\begin{itemize}
    \item Use ISO-8601 date format
    \item Maximum granularity: DAY
    \item Use \texttt{XXXX-XX-XX} if timestamps are unknown
\end{itemize}

\medskip

\textbf{Output Placeholder}
\begin{verbatim}
Timeline:
1    YYYY-MM-DD    docid-sentid-event
2    YYYY-MM-DD    docid-sentid-event
...
\end{verbatim}

\end{tcolorbox}

\paragraph{Entity-focused Configuration}\

\textbf{Query Objective:}

Construct timelines centered on events involving a target entity.

\textbf{Key Reasoning Focus:}
\begin{itemize}
    \item Entity mention and alias matching
    \item Argument extraction and participant consistency
    \item Entity coreference and bridging reasoning
\end{itemize}

\textbf{Representative Query Placeholder}

Target Entity:
{entity}

\paragraph{Aspect-focused Configuration}\

\textbf{Query Objective:}

Construct timelines describing different aspects of a crisis or complex event development.

\textbf{Key Reasoning Focus:}
\begin{itemize}
    \item Aspect semantic type classification
    \item Causal dependency reasoning
    \item Narrative-oriented event aggregation
\end{itemize}

\textbf{Representative Query Placeholder}

Aspect Query:
{aspect}

\paragraph{Seed-event-focused Configuration}\

\textbf{Query Objective:}

Construct timelines around events related to a given seed event.

\textbf{Key Reasoning Focus:}
\begin{itemize}
    \item Event semantic similarity
    \item Commonsense and causal relation reasoning
    \item Argument and temporal consistency modeling
\end{itemize}

\textbf{Representative Query Placeholder}

Seed Event:
{seed\_event}

\subsubsection{Implementation Details of TDAG}
\label{sec:c.3}

\begin{tcolorbox}[title=Autonomous Agent Prompt Template for QUEST, breakable]

\textbf{Task Objective}

Given a set of documents and a query, construct a query-focused event timeline by detecting relevant events, resolving cross-document event coreference, anchoring events in time, and ordering them chronologically.

\medskip

\textbf{Input Placeholder}
\begin{verbatim}
Documents:
{documents}

Query:
{query}

Query Type:
{query_type}
\end{verbatim}

\medskip

\textbf{Timeline Format}
\begin{verbatim}
ordering    time_anchor
docid-sentid-event
0           YYYY-MM-DD
docid-sentid-event
1           YYYY-MM-DD
docid-sentid-event
...
\end{verbatim}

\medskip

\hrule
\medskip

\textbf{Agent Reasoning Protocol}

You are an autonomous intelligent agent responsible for constructing the timeline.

\begin{itemize}
  \item Perform step-by-step reasoning.
  \item Issue only one action at a time.
  \item Use subtasks when complex reasoning is required.
\end{itemize}

Each response must follow:
\begin{verbatim}
THOUGHT: reasoning process
ACTION: selected action
\end{verbatim}

\medskip

\hrule
\medskip

\textbf{Available Actions}

\begin{itemize}
  \item \texttt{<action>subagent\_handle
  (subtask\_name)</action>}
  \item \texttt{<action>over()</action>}
\end{itemize}

\medskip

\textbf{Subtask Format}
\begin{verbatim}
<subtask>{
  "subtask_name": string,
  "goal": string,
  "criticism": string,
  "milestones": list[string],
  "result_format": optional
}</subtask>
\end{verbatim}

\medskip

\hrule
\medskip

\textbf{General Timeline Constraints}

\begin{itemize}
  \item Resolve cross-document event coreference conservatively
  \item Use ISO-8601 DATE format for timestamps
  \item Use \texttt{XXXX-XX-XX} if timestamps are unknown
  \item Order events using temporal relations
  \item Place temporally uncertain events at ordering index \texttt{0}
\end{itemize}

\medskip

\textbf{Final Output}

Output the final timeline within:
\begin{verbatim}
<timeline>
...
</timeline>
\end{verbatim}

Then call:
\begin{verbatim}
<action>over()</action>
\end{verbatim}

\end{tcolorbox}

\paragraph{Entity-focused Configuration}

\textbf{Query Placeholder}
\begin{verbatim}
Target Entity:
{entity}
\end{verbatim}

\textbf{Additional Constraints}
\begin{itemize}
    \item Include only events involving the target entity or its coreferential mentions
    \item Resolve entity-event participant associations
    \item Merge coreferential event mentions across documents
\end{itemize}

\paragraph{Aspect-focused Configuration}

\textbf{Query Placeholder}
\begin{verbatim}
Aspect Query:
{aspect}
\end{verbatim}

\textbf{Additional Constraints}
\begin{itemize}
    \item Focus on aspect-relevant crisis events
    \item Include causally related supporting events
    \item Construct coherent narrative-oriented timelines
\end{itemize}

\paragraph{Seed-event-focused Configuration}

\textbf{Query Placeholder}
\begin{verbatim}
Seed Event:
{seed_event}
\end{verbatim}

\textbf{Additional Constraints}
\begin{itemize}
    \item Identify semantically or causally related events
    \item Model argument and temporal consistency
    \item Preserve temporally associated event developments
\end{itemize}

\section{Benchmark Details}
\label{sec:appendix1}

\subsection{Aspect-focused Dataset Details}
\label{sec:appb.1}
Our crisis-oriented aspect schema consists of seven categories:
\textit{(1) Situation and Hazard},
\textit{(2) People Impact},
\textit{(3) Infrastructure and Lifeline},
\textit{(4) Needs and Help},
\textit{(5) Response and Recovery},
\textit{(6) Risk Communication}, and
\textit{(7) Source and Credibility}.

\paragraph{Crisis timeline construction.}
Starting from the EventStoryLine Corpus v1.5 annotations, we construct aspect-specific crisis timelines for each selected topic.
We first group documents by topic and extract event mentions together with their sentence and document identifiers.
For each crisis aspect, we select aspect-relevant seed events using aspect keyword rules, and then expand the seed set with causal predecessors annotated by \textsc{PlotLink}.
This gives each aspect timeline not only the target events of interest, but also the causal context leading to them.

We then anchor events to time using annotated \textsc{TLink} relations whenever the linked time expression can be normalized to a day-level ISO date.
Events that refer to the same action are merged using ACT cross-document coreference, while singleton events are kept as independent timeline nodes.
After temporal anchoring and ACT-based merging, events are sorted chronologically within each topic-aspect timeline.
Finally, we export the resulting timelines into a simple text format following prior timeline datasets: each line contains a timeline index, a normalized time value, and a compact event identifier.

\subsection{Illustrative Annotation Examples}
\label{sec:annotation_examples}

We present two representative annotation traces, one for each newly constructed query setting. Each trace follows the annotation stages from source evidence to the final cross-document timeline.

\subsubsection{Seed-event Query}

\textbf{Input:}

\noindent\textbf{Seed event:} \textit{Labour: We will introduce a Football Governance Bill, which will establish an independent regulator to ensure financial sustainability of football clubs in England}.\par
\noindent\textbf{Source documents:} 42 news articles.\par
\noindent\textbf{Existing annotations:} PledgeTracker relevant summarized event descriptions.

\paragraph{Step 1: Identify supporting sentences and event triggers.}
For each PledgeTracker event description, we additionally annotate its supporting sentence(s) in the source documents and the corresponding event trigger(s). Representative annotations are shown below.

{\footnotesize
\begin{center}
\setlength{\tabcolsep}{4pt}
\renewcommand{\arraystretch}{1.12}
\begin{tabular}{L{0.09\columnwidth}L{0.76\columnwidth}}
\toprule
\textbf{Doc.} & \textbf{Annotation} \\
\midrule
\textbf{6} & \textbf{Event description:} The bill to create the Independent Football Regulator had its second Reading in the House of Commons. \\
& \textbf{Source sentence:} \textit{Sent. 5: The bill to create the Independent Football Regulator had its second Reading in the House of Commons this week but amendments are likely to be added.} \\
& \textbf{Trigger(s):} \texttt{second\_Reading}, \texttt{added} \\
\addlinespace[3pt]
\textbf{8} & \textbf{Event description:} David Kogan, the government's preferred candidate for the independent regulator of English football, discusses the importance of the role in ensuring financial sustainability and the survival of the football pyramid. \\
& \textbf{Source sentence:} \textit{Sent. 4: Across two hours he set out his vision for a job he said had become ``more important and more pressing'' since it was first proposed and argued the regulator's job was ``to help the pyramid survive''.} \\
& \textbf{Trigger(s):} \texttt{set\_out}, \texttt{said}, \texttt{argued} \\
\addlinespace[3pt]
\textbf{12} & \textbf{Event description:} David Kogan, the government's preferred choice as chair of English football's new independent regulator, discusses the role and powers of the regulator during a pre-appointment hearing. \\
& \textbf{Source sentence:} \textit{Sent. 16: At a `pre-appointment hearing', Kogan---a former BBC journalist who also previously advised the Premier League, EFL and other leagues on broadcast rights---said he wants to put ``fans at the heart of the regulator'' and help the football pyramid ``survive''.} \\
& \textbf{Trigger(s):} \texttt{said} \\
\bottomrule
\end{tabular}
\end{center}
}

\paragraph{Step 2: Temporal annotation.}
Annotators normalize the time of each event trigger by combining explicit temporal expressions, document publication dates, and contextual temporal cues such as tense, aspect, and voice.

{\footnotesize
\begin{center}
\setlength{\tabcolsep}{3pt}
\renewcommand{\arraystretch}{1.12}
\begin{tabular}{L{0.40\columnwidth}L{0.18\columnwidth}L{0.29\columnwidth}}
\toprule
\textbf{Event trigger(s)} & \textbf{Time} & \textbf{Reasoning} \\
\midrule
\texttt{6-5-second\_Reading}, \texttt{6-5-added} & 2025-05-XX & Based on ``this week'' and the publication date, 2025-05-02. \\
\texttt{8-4-set\_out}, \texttt{8-4-said}, \texttt{8-4-argued} & 2025-05-07 & Based on the publication date, 2025-05-07, and the tense and voice of the source sentence. \\
\texttt{12-16-said} & 2025-05-07 & Based on the publication date, 2025-05-07, and the tense and voice of the source sentence. \\
\bottomrule
\end{tabular}
\end{center}
}

\paragraph{Step 3: Cross-document event coreference.}
After extracting events from multiple documents, annotators determine whether different event triggers refer to the same real-world event based on the trigger type and event arguments.

{\footnotesize
\begin{center}
\setlength{\tabcolsep}{3pt}
\renewcommand{\arraystretch}{1.12}
\begin{tabular}{L{0.32\columnwidth}L{0.28\columnwidth}L{0.26\columnwidth}}
\toprule
\textbf{Coreferential event triggers} & \textbf{Trigger type} & \textbf{Shared arguments} \\
\midrule
\texttt{8-4-argued}; \texttt{12-16-said} & Communication /\newline statement & Speaker: David Kogan; topic: the regulator's role; goal: helping the football pyramid survive \\
\bottomrule
\end{tabular}
\end{center}
}

\paragraph{Step 4: Timeline construction.}
Finally, event mentions are merged based on temporal information and cross-document event relations to construct the final timeline.

{\footnotesize
\begin{center}
\setlength{\tabcolsep}{4pt}
\renewcommand{\arraystretch}{1.08}
\begin{tabular}{L{0.11\columnwidth}L{0.22\columnwidth}L{0.51\columnwidth}}
\toprule
\textbf{Order} & \textbf{Time} & \textbf{Event} \\
\midrule
$\ldots$ & $\ldots$ & $\ldots$ \\
3 & 2025-05-XX & \texttt{6-5-second\_Reading} \\
3 & 2025-05-XX & \texttt{6-5-added} \\
$\ldots$ & $\ldots$ & $\ldots$ \\
5 & 2025-05-07 & \texttt{8-4-set\_out} \\
5 & 2025-05-07 & \texttt{8-4-said} \\
5 & 2025-05-07 & \texttt{8-4-argued}; \texttt{12-16-said} \\
$\ldots$ & $\ldots$ & $\ldots$ \\
\bottomrule
\end{tabular}
\end{center}
}

\subsubsection{Aspect Query}

\textbf{Input:}

\noindent\textbf{Aspect schema:} (1) Situation and Hazard, (2) People Impact, (3) Infrastructure and Lifeline, (4) Needs and Help, (5) Response and Recovery, (6) Risk Communication, and (7) Source and Credibility.\par
\noindent\textbf{Topic:} The Qeshm Island earthquake.\par
\noindent\textbf{Source documents:} 11 news articles.\par
\noindent\textbf{Existing annotations:} ECB+/ESC event mentions, event--time relations, intra-document causal relations, and cross-document event-coreference relations.

\paragraph{Step 1: Event type annotation.}
The source corpus provides the event triggers. We assign the existing event mentions to the seven analytical aspects used to construct aspect-specific timelines. In the examples below, italicized spans are source-annotated events, whereas the aspect in parentheses is the derived category in our benchmark.

{\footnotesize
\begin{center}
\setlength{\tabcolsep}{4pt}
\renewcommand{\arraystretch}{1.12}
\begin{tabular}{L{0.08\columnwidth}L{0.77\columnwidth}}
\toprule
\textbf{Doc.} & \textbf{Sentence} \\
\midrule
2 & Sent. 3: The sixth \textit{aftershock} (\textbf{Situation and Hazard}) \textit{shook} (\textbf{Situation and Hazard}) this southern Iranian island \ldots{} as \textit{death} (\textbf{People Impact}) toll from a strong \textit{earthquake} (\textbf{Situation and Hazard}) a day before rose to 10. \\
\addlinespace
11 & Sent. 5: Ten people were \textit{killed} (\textbf{People Impact}) and at least 50 \textit{injured} (\textbf{People Impact}) when an \textit{earthquake} (\textbf{Situation and Hazard}) razed mud-brick villages \ldots{} Sunday, officials and state media \textit{said} (\textbf{Source and Credibility}). \\
\addlinespace
1 & Sent. 3: A powerful \textit{earthquake} (\textbf{Situation and Hazard}) has \textit{hit} (\textbf{Situation and Hazard}) southern Iran, \textit{killing} (\textbf{People Impact}) at least 10 people and \textit{injuring} (\textbf{People Impact}) dozens more, officials \textit{say} (\textbf{Source and Credibility}). \\
\bottomrule
\end{tabular}
\end{center}
}

\paragraph{Step 2: Normalizing event time from temporal links.}
The source annotation links events to temporal expressions. We normalize the value already attached to the linked temporal expression; we do not infer a date from world knowledge.

{\footnotesize
\begin{center}
\setlength{\tabcolsep}{3pt}
\renewcommand{\arraystretch}{1.12}
\begin{tabular}{L{0.31\columnwidth}L{0.35\columnwidth}L{0.21\columnwidth}}
\toprule
\textbf{Event} & \textbf{Existing temporal link} & \textbf{Time} \\
\midrule
\texttt{2-3-aftershock} & \texttt{11/28/05 CONTAINS aftershock} & 2005-11-28 \\
\texttt{2-3-shook} & \texttt{11/28/05 CONTAINS shook} & 2005-11-28 \\
\texttt{9-1-hits} & No usable event--time relation & \texttt{XXXX-XX-XX} (unknown) \\
\bottomrule
\end{tabular}
\end{center}
}

\paragraph{Step 3: Document-level timeline construction.}
For each article, we first select the events belonging to the requested aspect. We then add only predecessors licensed by an existing intra-document causal relation and consolidate triggers belonging to the same event. Representative causal relations are shown below.

{\footnotesize
\begin{center}
\setlength{\tabcolsep}{3pt}
\renewcommand{\arraystretch}{1.1}
\begin{tabular}{L{0.08\columnwidth}L{0.21\columnwidth}L{0.20\columnwidth}L{0.31\columnwidth}}
\toprule
\textbf{Doc.} & \textbf{Predecessor} & \textbf{Aspect event} & \textbf{Causal relation} \\
\midrule
2 & earthquake & aftershocks & \textsc{falling\_action} \\
11 & earthquake & aftershocks & \textsc{precondition} \\
11 & earthquake & hit & \textsc{precondition} \\
1 & quake & struck & \textsc{precondition} \\
1 & tremor & aftershocks & \textsc{precondition} \\
\bottomrule
\end{tabular}
\end{center}
}

\noindent\textit{Document-level timeline for Doc.~2.}
{\footnotesize
\begin{center}
\setlength{\tabcolsep}{4pt}
\renewcommand{\arraystretch}{1.08}
\begin{tabular}{L{0.11\columnwidth}L{0.22\columnwidth}L{0.51\columnwidth}}
\toprule
\textbf{Order} & \textbf{Time} & \textbf{Event} \\
\midrule
1 & 2005-11-27 & \texttt{2-3-earthquake}; \texttt{2-4-quake} \\
1 & 2005-11-27 & \texttt{2-17-quake}; \texttt{2-18-earthquake} \\
1 & 2005-11-27 & \texttt{2-4-hit}; \texttt{2-16-hit} \\
2 & 2005-11-28 & \texttt{2-3-aftershock} \\
$\ldots$ & $\ldots$ & $\ldots$ \\
\bottomrule
\end{tabular}
\end{center}
}

\noindent\textit{Document-level timeline for Doc.~11.}
{\footnotesize
\begin{center}
\setlength{\tabcolsep}{4pt}
\renewcommand{\arraystretch}{1.08}
\begin{tabular}{L{0.11\columnwidth}L{0.22\columnwidth}L{0.51\columnwidth}}
\toprule
\textbf{Order} & \textbf{Time} & \textbf{Event} \\
\midrule
1 & 2005-11-27 & \texttt{11-1-Earthquake}; \texttt{11-5-earthquake} \\
2 & 2005-11-27 & \texttt{11-11-aftershocks} \\
2 & 2005-11-27 & \texttt{11-16-hit} \\
$\ldots$ & $\ldots$ & $\ldots$ \\
\bottomrule
\end{tabular}
\end{center}
}

\paragraph{Step 4: Global timeline construction.}
We next merge document-level nodes only when the source annotation places their mentions in the same cross-document event-coreference group.

{\footnotesize
\begin{center}
\setlength{\tabcolsep}{3pt}
\renewcommand{\arraystretch}{1.12}
\begin{tabular}{L{0.38\columnwidth}L{0.45\columnwidth}}
\toprule
\textbf{Source coreference group} & \textbf{Coreferential event triggers} \\
\midrule
{\scriptsize\texttt{ACT17821472137518203}} & \texttt{2-3-earthquake}; \texttt{11-5-earthquake}; \texttt{1-3-earthquake} \\
{\scriptsize\texttt{ACT17822303993286740}} & \texttt{2-4-hit}; \texttt{11-16-hit}; \texttt{1-3-hit} \\
{\scriptsize\texttt{ACT17821815048905207}} & \texttt{2-2-Aftershocks}; \texttt{11-11-aftershocks}; \texttt{1-15-aftershocks} \\
\bottomrule
\end{tabular}
\end{center}
}

\noindent\textit{Resulting global timeline.}
{\footnotesize
\begin{center}
\setlength{\tabcolsep}{4pt}
\renewcommand{\arraystretch}{1.08}
\begin{tabular}{L{0.11\columnwidth}L{0.22\columnwidth}L{0.51\columnwidth}}
\toprule
\textbf{Order} & \textbf{Time} & \textbf{Event} \\
\midrule
0 & \texttt{XXXX-XX-XX} & \texttt{9-1-hits} \\
$\ldots$ & $\ldots$ & $\ldots$ \\
1 & 2005-11-27 & \texttt{2-3-earthquake}; \texttt{11-5-earthquake}; \texttt{1-3-earthquake} \\
1 & 2005-11-27 & \texttt{2-4-hit}; \texttt{11-16-hit}; \texttt{1-3-hit}; \texttt{1-4-struck} \\
1 & 2005-11-27 & \texttt{2-2-Aftershocks}; \texttt{11-11-aftershocks}; \texttt{1-15-aftershocks} \\
$\ldots$ & $\ldots$ & $\ldots$ \\
2 & 2005-11-28 & \texttt{2-3-aftershock} \\
2 & 2005-11-28 & \texttt{2-3-shook} \\
$\ldots$ & $\ldots$ & $\ldots$ \\
\bottomrule
\end{tabular}
\end{center}
}

\paragraph{Step 5: Topic-level quality review.}
After constructing the candidate aspect timelines, we conduct a topic-level quality review to ensure the reliability of the resulting benchmark. We mainly remove topics whose aspect timelines are too sparse to support meaningful timeline construction, as well as topics where cross-document coreference introduces many conflicts with the temporal and causal relations.

\subsection{Evaluation Metric}
\label{sec:appb.2}
We adopt the official evaluation protocol of the SemEval-2015 TimeLine task\footnote{https://alt.qcri.org/semeval2015/task4/}, which is based on the TempEval-3 metric~\cite{uzzaman-etal-2013-semeval}. This metric comprehensively measures system performance by assessing the accuracy of events, temporal expressions, and their interrelations. Final scores are averaged across all timelines in the dataset.

\paragraph{Event correctness.}
Event correctness is directly reflected through the graph matching process. Since event mentions constitute the nodes of the temporal graph, incorrect or spurious events cannot be aligned with gold event nodes, causing their associated temporal links to reduce precision. Conversely, missing relevant events leads to missing graph structures and decreases recall.

\paragraph{Temporal grounding.}
The time component of each $(\text{time}, \text{event})$ entry determines both the event's normalized anchor and its relative position in the timeline. An incorrect anchor changes the event--time relation and the event--event relations induced by temporal closure. The graph comparison therefore evaluates whether selected events are grounded and ordered consistently with the gold timeline.

\paragraph{Cross-document event coreference.}
Cross-document event organization is also reflected in the temporal graph structure. Events originating from different documents are evaluated based on their positions within the timeline. Incorrectly merging, separating, or aligning cross-document event mentions changes the resulting graph structure, leading to lower graph precision and recall.

\section{Experimental Details}
\label{sec:appendix3}

\subsection{SFT Hyperparameter Settings}
\label{sec:appb.3}

To alleviate the difficulty of directly learning long-range cross-document reasoning under extremely limited supervision, we first decompose gold timelines into document-level timelines for single-document training, and then further train on the original cross-document timelines. Detailed training configurations are shown in Table~\ref{tab:lora_config}.

\begin{table}[t]
\centering
\small
\begin{tabular}{ll}
\hline
\textbf{Parameter} & \textbf{Value} \\ \hline
Train Type & LoRA \\
LoRA Rank ($r$) & 16 \\
LoRA Alpha & 32 \\
Target Modules & all-linear  \\
Learning Rate & $2\times10^{-4}$ \\
Warmup Ratio & 0.05 \\
Drop & 0.05 \\
Batch Size & 4 \\
Epochs & 10 \\
Random Seeds & 3 (results averaged) \\
\hline
\end{tabular}
\caption{Training configuration for LoRA-based SFT experiments.}
\label{tab:lora_config}
\end{table}

\subsection{Qualitative Error Analysis}
\label{sec:app3.3}

As shown in Table~\ref{tab:error_prop}, Detection, Matching, \& Filtering plays the most fundamental role across all query types. This stage is also the starting point of the pipeline: if the relevant events are incorrectly detected or matched at the beginning, later temporal anchoring and cross-document coreference may still be correct locally, but the final timeline will remain incorrect because it is built on the wrong event set. We give qualitative error analysis to make this failure mode clearer, organized by query type:

\paragraph{Entity query.}
Examples illustrate errors in matching target entities to the correct
events, particularly when event arguments are related but not strictly
coreferential.

\begin{itemize}
    \item \textbf{Sentence 1:} The 16GB version is \textbf{priced} at US dollar 199 and the 32GB version at US dollar 299.

    \item \textbf{Analysis:} ``The 16GB version'' is an argument of the
    \textbf{priced} event and is related to iPhone 4 through a
    product-variant relation, but it is not coreferential with the target
    entity iPhone 4.

    \item \textbf{Sentence 2:} On September 21, 2004 the parties
    \textbf{agreed} to have the case heard by the UK court.

    \item \textbf{Analysis:} ``The parties'' is an argument of the
    \textbf{agreed} event and collectively refers to ``Apple Inc.'' and
    ``Apple Corps'', but it does not individually corefer with either
    company.
\end{itemize}

\paragraph{Aspect query.}
The query requires identifying events matching the requested aspect
(human impact) and connecting them to earlier causal or background events
in the document.

\begin{itemize}
    \item \textbf{Aspect-relevant sentence:} Comeaux also \textbf{injured}
    a man who was visiting another inmate at the time and tried to stop
    the \textbf{attack}.

    \item \textbf{Classification:} \textbf{injured} and \textbf{attack}
    are classified as People Impact events under the seven-aspect schema.

    \item \textbf{Earlier causal/background event:} He was serving a life
    sentence after being convicted in June 1998 of aggravated sexual
    \textbf{assault} out of Brazos County, located northwest of Houston.

    \item \textbf{Analysis:} Beyond correctly classifying \textbf{injured}
    and \textbf{attack}, the query requires linking them to the earlier
    \textbf{assault} event, which provides the relevant causal or
    background context.
\end{itemize}

\paragraph{Seed-event query.}
The query involves two complementary judgments: (1) filtering out events
that share topical or lexical overlap with the seed event but do not
correspond to the pledge, and (2) identifying events expressed with
different wording that nevertheless provide evidence about the pledge
status.

\begin{enumerate}
    \item \textbf{Filtering out lexical overlap without pledge relevance}

    \begin{itemize}
        \item \textbf{Seed event:} Labour: We will support families with
        children by introducing free breakfast clubs in every primary
        school.

        \item \textbf{Sentence event:} Twenty five schools in Devon were
        among the first in the country to offer free, daily breakfast
        clubs.

        \item \textbf{Analysis:} Although the sentence mentions breakfast
        clubs, it describes a local implementation in a limited set of
        schools, whereas the seed event concerns a national pledge to
        introduce free breakfast clubs in every primary school. This
        requires distinguishing local evidence from the global policy
        commitment.
    \end{itemize}

    \item \textbf{Identifying implicit evidence with different wording}

    \begin{itemize}
        \item \textbf{Seed event:} Labour: Labour will not increase taxes
        on working people, which is why we will not increase National
        Insurance.

        \item \textbf{Sentence event:} Direct taxes as a percentage of
        gross income for all UK households remained relatively stable at
        22.7\%, between FYE 2023 and FYE 2024.

        \item \textbf{Analysis:} The sentence does not explicitly mention
        ``National Insurance'' or ``working people'', but the share of
        direct taxes in household income provides evidence about whether
        the tax burden increased, making the event relevant to the pledge.
    \end{itemize}
\end{enumerate}

\subsection{Case Study}
\label{sec:app3.2}

We first examine a factuality driven example:
\emph{``Neither company made any official statement about the technical issues.''}
In this case, the light verb \emph{made} serves as syntactic support and does not independently introduce an event.
Instead, the nominal predicate \textbf{statement}\textsubscript{NegatedFactual} acts as the event trigger, whose factuality is determined by the scope of syntactic negation and filtered out.

We further present a representative case where the event mention \textit{offers} in ``A primary school offers a Breakfast Club running from 7:45am every day'' should not be considered related to the seed event ``Labour: We will support families with children by introducing free breakfast clubs in every primary school,'' because their event participants and affected scopes are fundamentally different. The former describes a routine activity conducted by an individual school, while the latter refers to a nationwide policy initiative proposed by a political organization.

\end{document}